\documentclass[10pt, conference, compsocconf]{IEEEtran}

\usepackage{algorithm}
\usepackage{algpseudocode}

\usepackage{amsmath}
\usepackage{graphicx}

\usepackage{epstopdf}

\usepackage{bbm}
\usepackage{amsthm} 
\usepackage{amssymb}

\usepackage{algpseudocode}
\usepackage{algorithm}

\usepackage[backref=page]{hyperref}

\hypersetup{
    colorlinks = true, 
    linkcolor = blue,
    anchorcolor = red,
    citecolor = blue, 
    filecolor = red, 
    pagecolor = red,
    urlcolor = red,
}

\usepackage{amsfonts, amsbsy, amssymb}

\usepackage[numbers]{natbib}

\begin{document}
%
\title{Inverse Kinematics with Dual-Quaternions,\\ Exponential-Maps, and Joint Limits}


\author{\IEEEauthorblockN{Ben Kenwright}
\IEEEauthorblockA{	Newcastle University\\
					School of Computing Science\\
					United Kingdom\\
					b.kenwright@ncl.ac.uk}
}

\maketitle

\def\mykeywords{Inverse Kinematics; Gauss-Seidel; Articulated Character; Games; Joint Limits; Iterative; Dual-Quaternion; Jacobian; Exponential-Map}

\begin{abstract}
We present a novel approach for solving articulated inverse kinematic problems (e.g., character structures) by means of an iterative dual-quaternion and exponential-mapping approach.  As dual-quaternions are a break from the norm and offer a straightforward and computationally efficient technique for representing kinematic transforms (i.e., position and translation).  Dual-quaternions are capable of represent both translation and rotation in a unified state space variable with its own set of algebraic equations for concatenation and manipulation.  Hence, an articulated structure can be represented by a set of dual-quaternion transforms, which we can manipulate using inverse kinematics (IK) to accomplish specific goals (e.g., moving end-effectors towards targets).  We use the projected Gauss-Seidel iterative method to solve the IK problem with joint limits.  Our approach is flexible and robust enough for use in interactive applications, such as games.  We use numerical examples to demonstrate our approach, which performed successfully in all our test cases and produced pleasing visual results.  
\end{abstract}


%

\section{Introduction}
Generating fast reliable Inverse Kinematic (IK) solutions in real-time with angular limits for highly articulated figures (e.g., human bipeds including hands and feet) is challenging and important \cite{Ken12a,Ken12b,Welman1993,Unzueta2008,Wolovich1984}.  The subject is studied across numerous disciplines, such as graphics, robotics, and biomechanics, and is employed by numerous applications in the film, animation, virtual reality, and game industry

However, articulated models (e.g., bipeds and hands) can be highly complex; even the most simplified models of 20-30 joints can generate a vast number of poses \cite{Kulpa2005,Zhao1994}.  Whereby producing a simple pose to achieve a solitary task can produce ambiguous solutions that make the problem highly nonlinear and computationally expensive to solve.  For example, even a straightforward task of reaching to pickup an object can be accomplished by means of any number of motions. 

This paper focuses using dual-quaternions and quaternion exponential-maps with an iterative Gauss-Seidel algorithm \cite{Cottle1992} to solve an articulated IK problem; such as the hand model shown in Figure \ref{fig:articulated_model}.  The Gauss-Seidel algorithm is an iterative, efficient, low memory method of solving linear systems of equations of the form $Ax=b$.  Hence, we integrate the Gauss-Seidel iterative algorithm with an articulated IK problem to produce a flexible whole system IK solution for time critical systems, such as games.  This method is used as it offers a flexible, robust solution with the ability to trade accuracy for speed and give good visual outcomes. 

Furthermore, to make the Gauss-Seidel method a practical IK solution for an articulated hand structure, it needs to enforce joint limits.  We incorporate joint limits by modifying the update scheme to include an iterative projection technique.  Additionally, to ensure real-time speeds we take advantage of spatial coherency between frames as a warm starting approximation for the solver.  Another important advantage of the proposed method is the simplicity of the algorithm and how it can be easily configured for custom IK problems. 

The main contribution of the paper is the practical demonstration and discussion of using the Gauss-Seidel method for real-time articulated IK problem with joint limits, dual-quaternions \cite{Ken12c}, and exponential-quaternion mapping \cite{Gras98}.  Furthermore, we discuss constraint conditions, speedup approaches and robustness factors for solving highly non-linear IK problems in real-time.

The roadmap for rest of the paper is organized as follows.  Firstly, we briefly review existing work in Section \ref{sec:related_work}.  Section \ref{sec:articulated_model} describes the articulated model, we use for our simulations.  Then in Section \ref{sec:math_background}, we present essential mathematical algorithms and principles for the paper (e.g., dual-quaternion algebra).  We follow on by explaining the IK problem in Section \ref{sec:fk_ik}. While in Section \ref{sec:jacobian_matrix}, we explain the Jacobian matrix, then in Section \ref{sec:gauss_seidel_algorithm} we discuss our approach for solving the IK problem with the Gauss-Seidel algorithm.  Finally, we present results in Section \ref{sec:experimental_results}, then Section \ref{sec:limitations} discusses limitations, followed by the closing conclusion and discussion in Section \ref{sec:conclusion_and_discussion}.

\begin{figure}
	\begin{center}
	\includegraphics[width=0.9\columnwidth]{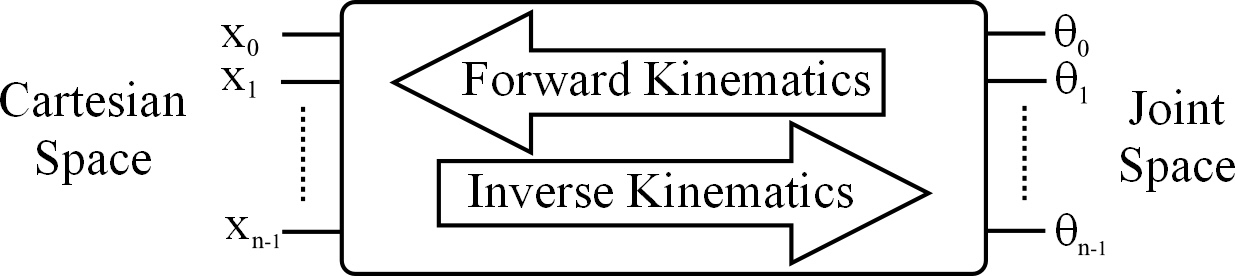}
	\end{center}
	\caption{\text{Forward \& Inverse Kinematics}.  Illustrating the relationship between forward and inverse kinematics parameters.}
	\label{fig:understanding_ik}
\end{figure}

\section{Related Work} \label{sec:related_work}
Inverse kinematics is a popular problem across numerous disciplines (e.g., graphical animation, robotics, biomechanics).  IK is a vital component that can be implemented using a wide range of solutions.  We give a brief overview of existing, current, and cutting-edge approaches to help emphasise the different ways of approaching the problem; enabling the reader to see where our method sits. 

In general, however, for very simple problems with just a few links, analytical methods are employed to solve the IK problem.  Alternatively, for larger configurations, iterative numerical methods must be employed due to the complexity of the problem.

The articulated IK problem of finding a solution for poses that satisfy positional and orientation constraints has been well studied, e.g., \cite{Girard1985,Welman1993,Rose1997,Zhao1994}.  The problem is highly nonlinear, meaning there can be numerous solutions; hence, multiple poses fulfilling the constraint conditions.  In practical situations, there can even be cases where no solution exists due to the poor placement of end-effectors.  IK systems typically use cut down models, e.g., merely performing IK on individual limbs (as in body, arms, legs) \cite{Yamane2003,Zhao1998,Kulpa2005}.  This makes the problem computationally simpler and less ambiguous.

Numerous solutions from various fields of research have been implemented to solving the IK problem.  The Jacobian-based matrix approach is one of the most popular methods and the method upon which we base our iterative solution \cite{Zhao1994,Buss2004,Fletcher1987}.  The Jacobian matrix method aims to find a linear approximation to the problem by modelling the end-effectors movements relative to the instantaneous systems changes of the links translations and orientations.  Numerous different methods have been presented for calculating the Jacobian inverse, such as, Jacobian Transpose, Damped Least-Squares (DLS), Damped Least-Squares with Singular Value Decomposition (SVD-DLS), Selectively Damped Least-Square (SDLS) \cite{Balestrino1984,Wolovich1984,Baillieul1985,Wampler1986,Nakamura1986,Buss2004}.

An alternative method uses the Newton method; whereby the problem is formulated as a minimization problem from which configuration poses are sought.  The method has the disadvantage of being complex, difficult to implement and computationally expensive to computer-per-iteration \cite{Fletcher1987}.

The Cyclic Coordinate Descent (CCD) is a popular real-time IK method used in the computer games industry \cite{Lander1998}.  Originally introduced by Wang et al. \cite{Wang1991} and then later extended to include constraints by Welman et al. \cite{Welman1993}.  The CCD method was designed to handle serial chains and is thus difficult to extend to complex hierarchies.  It has the advantage of not needing to formulate any matrices and has a lower computational cost for each joint per iteration.  Its downside is that the character poses even with constraints can produce sporadic and unrealistic poses.  However, further work has been done to extend CCD to work better with human based character hierarchies \cite{Unzueta2008,Kulpa2005,Boulic2006}.

A novel method recently proposed was to use a Sequential Monte Carlo approach but was found to be computationally expensive and only applicable for offline processing \cite{Courty2008,Hecker2008}. 

Data driven IK systems have been presented; Grochow et al. \cite{Grochow2004} method searched a library of poses to determine an initial best guess solution to achieve real-time results.  An offline mesh-based for human and non-human animations was achieved by learning the deformation space; generating new shapes while respecting the models constrains \cite{Sumner2005,Der2006}.

A method known as "Follow-The-Leader" (FTL) was presented by Brown et al. \cite{Brown2004} and offered real-time results using a non-iterative technique.  However, this approach was later built upon by Aristidou et al. \cite{Aristidou2011} and presented an iterative version of the solver known as FABRIK.

The Triangular IK method \cite{Mukundan2009,Muller-Cajar2007}, uses trigonometric properties of the cosine rule to calculate joint angles, beginning at the root and moving outwards towards the end-effectors.  While the algorithm can be computationally fast, due to it being able to propagate the full hierarchy in a single iteration, it cannot handle multiple end-effectors well and is primarily based around singly linked systems.

The advantages of an iterative IK system for articulated structures, such as character, was also presented by the interesting paper by Tang et al. \cite{Tang1999} who explored IK techniques for animation using a method based on the SHAKE algorithm.  The SHAKE algorithm is an iterative numerical integration scheme considered similar to the Verlet method \cite{Verlet1967}, which can exploit substantial step-sizes to improve speed yet remain stable when solving large constrained systems.  The algorithm is also proven to have the same local convergence criterion as the Gauss-Seidel method we present here as long as the displacement size is kept sufficiently small.

The paper by Arechavaleta et al. \cite{ALM09} presents a well written explanation of using iterative methods (primarily the Gauss-Seidel technique) for computing fast and accurate solutions for ill-conditioned LCP problems.

We use the iterative Gauss-Seidel approach presented by Kenwright \cite{Ken12a}, however, we store the joint angles as quaternion-exponent and employ dual-quaternion algebra \cite{Ken12c} for solving forward and inverse kinematic problems.  As unit-quaternions are an ideal tool for orientations of rigid transforms, however, they do not contain any translation information about the location of points in 3D space.  Hence, dual-quaternions are an extension of quaternion by means of dual-number theory as a compact, efficient, and smart approach of representing both rotation and translation in a single vector with its own algebraic rules (e.g., calculating differences, inverting, interpolating).  The robotic community has exploited dual-quaternions to solve IK problems; for example, Hoang-Lan et al. \cite{PPAF10} solved IK problem by formulating a dual-quaternion metric error measurement for constructing the Jacobian.

\section{Articulated Model} \label{sec:articulated_model}
\begin{figure}
	\begin{center}
	\includegraphics[width=0.9\columnwidth]{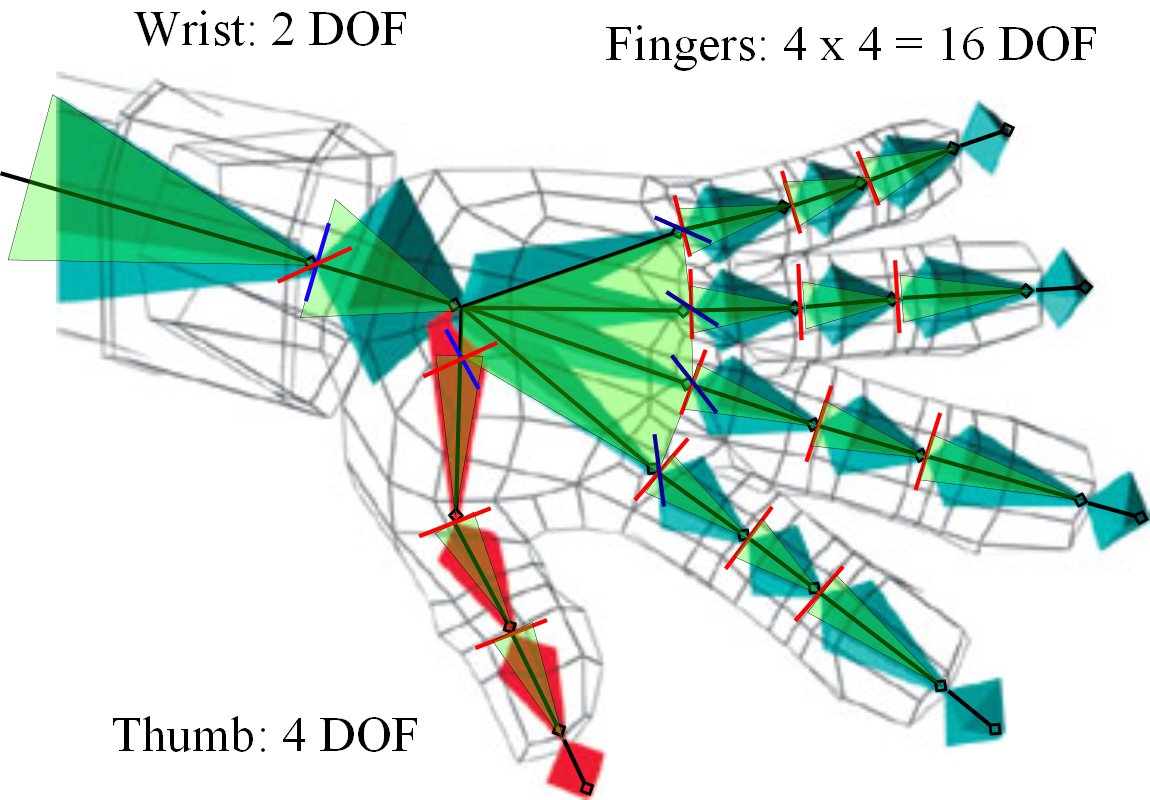}
	\end{center}
	\caption{\text{Articulated Hand Model}.  The articulated hand model (including the wrist) was used for the simulations and possessed 22 degrees of freedom (DOF).  The structure comprised of 17 links and 16 joints.}
	\label{fig:articulated_model}
\end{figure}

The articulated model used for our simulations and evaluation of our approach is shown in Figure \ref{fig:articulated_model}.  The mechanical functioning is of a human hand and is constructed using a series of interconnected rigid segments (or links) connected by joints (also, note, an interconnected series of links is also called a kinematic chain).  As shown in Figure \ref{fig:articulated_model}, we represent the hand as a collection of 17 rigid body segments connected using 16 primary joints.  The character's hand gives us 22 degrees of freedom (DOF).  Joints such as the wrist have three DOF corresponding to abduction/adduction, flexion/extension and internal/external rotation (i.e., rotation around the x, y, and z axis).  Where complex joints such as the ball-and-socket (i.e., with 3 DOF) can be formed from multiple simpler joints (i.e., 3 single DOF joints).  So a joint with n DOF is equivalent to n joints of 1 DOF connected by n-1 links of length zero.  Thus, for example, the wrist joint can be described as a sequence of 2 separate joints of 1 DOF, where 1 of the joints connecting links has a zero length, as done by Kenwright \cite{Ken12a}.  Euler angles are an intuitive and straightforward means of representing orientation, since they are easy to visualize and enforce upper and lower boundary limits.  However, we store each joint rotation as a quaternion-exponential (i.e., axis-angle combination), as it offers a similarly compact parameterization as Euler angles but without the gimbals lock problem.  We combine the exponential-mapping with quaternion and dual-quaternion algebra to solve the IK solution, while clamping angular limits through a twist-and-swing decomposition of the orientation.

As shown in Figure \ref{fig:articulated_model}, the single DOF connected joints were colored in accordance with their axis type; the x, y, and z representing the colors red, green and blue.  The foot was set as the base for the IK with five end-effectors (i.e., head, pelvis, right-hand, left-hand and left-foot).  We developed an application for an artist to interrogate and experiment with the skeletal IK system; setting end-effectors locations and viewing the generated poses.  Each end-effector has a 6 DOF constraint applied to it; representing the target position and orientation.  The ideal end-effectors are drawn in red, and the current end-effectors are drawn in green.  This can be seen clearly in Figure 5, where the target end-effectors are located at unreachable goals.

\section{Mathematical Background} \label{sec:math_background}
We give a brief introduction to the essential mathematical definitions on quaternion and dual-quaternion algebra.  For a more detailed introduction and an overview of their practical advantages, see Kenwright \cite{Ken12c}.
Since quaternions have proven themselves in many fields of science 
and computing as providing an unambiguous, un-cumbersome, computationally efficient method of representing rotational information.  We combine dual-number theory to extend quaternions to dual-quaternions, so we can use them to represent rigid transforms (i.e., translations and rotations).

\subsection{Definitions}
To reduce ambiguity and make the paper as readable as possible, we define variable symbol definitions:
\begin{equation}
\begin{alignedat}{3}
q \text{ quaternion}				  \quad  & \hat{q} \text{ unit-quaternion}\\
\underline{q} \text{ dual-quaternion} \quad  & \underline{\hat{q}} \text{ unit dual-quaternion} \\
\vec{v} \text{ vector } 			  \quad  & \hat{v} \text{ unit-vector} \\
\end{alignedat}
\end{equation}
While we mostly represent quaternions and dual-quaternions with the letter $q$, there are instances where we use the letter $t$ to indicate that the dual-quaternion or quaternion is a pure translation component.

\subsection{Quaternions}
The quaternion was introduced by Hamilton \cite{Ham60} in 1860 and has the following form:
\begin{equation}
	\textbf{q} = q_s + q_x\textbf{i} + q_y\textbf{j} + q_z\textbf{k} \quad (q_s,q_x,q_y,q_z \in \mathbb{R})
\end{equation}
\noindent where $\textbf{i}\textbf{i}=\textbf{j}\textbf{j}=\textbf{k}\textbf{k}=\textbf{i}\textbf{j}\textbf{k}=-1$.

While $w$ is sometimes used to represent the scalar component in quaternions, we us the letter $s$, to avoid ambiguity with exponential-mapping variable ($\vec{w}$).
Alternatively, it is more commonly defined as a pair $(s,\vec{v})$ with $s \in \mathbb{R}$ and $\vec{v} \in \mathbb{R}^3$.

A \textit{unit-quaternion} has a unit-length with $q_s^2+q_x^2+q_y^2+q_z^2=1$ and the inverse of a unit-quaternion is its conjugate $q^*=q_s - \vec{v}$.  Given an axis and angle of rotation a unit-quaternion can be calculated using:
\begin{equation}
\begin{alignedat}{3}
	q_s &= cos(\theta/2),	    &\quad  q_x &= n_x sin(\theta/2)\\
	q_y &= n_y sin(\theta/2),	&\quad  q_z &= n_z sin(\theta/2)
\end{alignedat}
\end{equation}

\noindent where $\hat{n}={n_x,n_y,n_z}$ is a unit-vector representing the axis of rotation and $\theta$ is the angle of rotation.  Whereby, given a point in 3D space $\vec{x}$ we can rotate to give $\vec{x}'$ using:
\begin{equation}
	\vec{x}'=\hat{q} \vec{x} \hat{q}^*
\end{equation}
Addition, subtraction, and the product of two quaternions, is defined by:
\begin{equation}
\begin{alignedat}{1}
	{\bf q_0} + {\bf q_1} &= (s_0+s_1, \vec{v_0} + \vec{v_1})\\
	{\bf q_0} - {\bf q_1} &= (s_0-s_1, \vec{v_0} - \vec{v_1})\\
	{\bf q_0}{\bf q_1}    &= (s_0 s_1 - \vec{v_0}\vec{v_1},
								  s_0\vec{v_1} + s_1\vec{v_0} + \vec{v_0}\vec{v_1})\\
\end{alignedat}
\end{equation}
A crucial fact that we exploit in this paper is that the exponential of a unit-quaternion is the combined axis-angle component:

\begin{equation}
\begin{alignedat}{3}
	exp( \hat{q} ) &= [0, \hat{n}\theta]\\
				   &= [0, \vec{w}]
\end{alignedat}
\end{equation}

\noindent where $\hat{n}$ is the unit-vector representing the axis of rotation and $\theta$ is the angle magnitude in radians.  The logarithm of a quaternion is the inverse of the exponential enabling us to convert to and from the axis-angle component.  The exponential of a unit-quaternion is often called the \textit{exponential map}, which we denote as $\vec{w}$.  

\begin{equation}
\begin{alignedat}{3}
	\theta  &= ||\vec{w}||\\
	\hat{v} &= \frac{\vec{w}}{||\vec{w}||}
\end{alignedat}
\end{equation}

The exponential-map can be computed robustly, even in the neighborhood of the origin \cite{Gras98}.

\subsection{Dual-Numbers}
Dual-number theory was introduced by Clifford \cite{Clif73} in 1873 and is defined as:
\begin{equation}
\begin{alignedat}{2}
	\hat{z} = r + \epsilon d & \quad \text{with } \epsilon^2=0 \text{ but } \epsilon \ne 0
\end{alignedat}
\end{equation}

\noindent where $r$ is the real-part, $d$ is the dual-part, and $\epsilon$ is the dual operator.
While dual-number theory can be used to represent different quantities (e.g., dual-vectors), we are primarily interested in dual-quaternions because it gives us the ability of unifying rigid transform space into a single state-space variable (i.e., position and translation).

\subsection{Dual-Quaternions}
A dual-quaternion is defined as a dual-number with quaternion components and has the ability to represent 3D Euclidean coordinate space (i.e., rotation and translation) as a single parameter.
\begin{equation}
	\underline{q} = q_r + \epsilon q_d
\end{equation}

\noindent where $q_r$ and $q_d$ are quaternions.  Additionally, since the dual-quaternion consists of two quaternion components it can be represented as: $\underline{q}=[q_0,q_1,q_2,q_3,q_4,q_5,q_6,q_7]^T$.
The common algebraic operations are defined as:
\begin{equation}
\begin{alignedat}{3}
	\alpha \underline{q} &= \alpha q_r + \alpha \epsilon q_d\\
	\underline{q}_0 + \underline{q}_1 &= q_{r0} + q_{r1} + 
										 \epsilon(q_{d0} + q_{d1})\\
	\underline{q}_0 \, \underline{q}_1 &= q_{r0}q_{r1} + 
									  \epsilon(q_{r0}q_{d1}+q_{d0}q_{r1})
\end{alignedat}
\end{equation}
The conjugate of a dual-quaternion $\underline{q}^* = q_r^* + \epsilon q_d^*$ with the norm (or length) of a dual-quaternion given by $||\underline{q}|| = \underline{q}\underline{q}^*$ and the unity condition (i.e., for unit dual-quaternions) is:

\begin{equation}
\begin{alignedat}{2}
	\underline{q} \, \underline{q}^* = 1 \quad q_r^* q_d + q_d^* q_r = 0
\end{alignedat}
\end{equation}

A \textbf{unit dual-quaternion} can be used to represent any rigid transformation (i.e., position and rotation); we construct a unit dual-quaternion rigid transformation using:

\begin{equation} \label{eq:dual_quat_parts}
\begin{alignedat}{2}
	\underline{\hat{q}} &= q_{rot} + \epsilon \frac{1}{2} q_{rot} q_{pos} 
	\quad && \text{(rotation then translation)}\\
	&&\qquad \text{or}\\
	\underline{\hat{q}} &= q_{rot} + \epsilon \frac{1}{2} q_{pos} q_{rot}
	\quad && \text{(translation then rotation)}\\
					    & = (1 + \epsilon \frac{1}{2} q_{pos})q_{rot}  && \\
\end{alignedat}
\end{equation}

\noindent where $q_{rot}$ and $q_{pos}$ are the rotation and translation quaternions respectively, with the translation quaternion $q_{pos}=[0,t_x,t_y,t_z]$.


\section{Forward and Inverse Kinematics} \label{sec:fk_ik}
Forward and inverse kinematics is the process of calculating positions and orientations either from joint space (i.e., using interconnected positions and orientations of the joints) or from Cartesian space (i.e., the world positions and orientations) as shown in Figure \ref{fig:understanding_ik}.


\subsection{Forward Kinematics (FK)}  The FK problem is straightforward to calculate and has no ambiguity or singularities.  For an articulated structure, we can concatenate the dual-quaternion transforms through multiplication to generate the final positions and orientations of the interconnected links.  For example, a serial chain of $n$ links would be:
\begin{equation} \label{eq:forward_kinematics}
	\underline{\hat{q}} = 	\underline{\hat{q}}_0
							\underline{\hat{q}}_1
							\underline{\hat{q}}_2
							...
							\underline{\hat{q}}_{n-1}
\end{equation}

\noindent where $\underline{\hat{q}}_0 ... \underline{\hat{q}}_{n-1}$ define each individual joints rotation and translation.


\subsection{Inverse Kinematics (IK)}  IK is the reverse of FK.  While FK remains fast and simple for large interconnected structures, IK solutions can be computationally expensive, possess singularity problems and contain multiple solutions.  However, in practice, we attempt to find a best fit approximation that will meet the desired constraints.  For example, in Equation (\ref{eq:forward_kinematics}), we know $\underline{\hat{q}}$ and would seek to find the orientation (and/or translation) for each link ($\underline{\hat{q}}_0 .. \underline{\hat{q}}_{n-1}$), and $\underline{\hat{q}}$ is the position and orientation of the combined links (i.e., for the end-effector).


\subsection{Pure Rotation and Pure Translation}
A dual-quaternion's transformation can be represented by two pure dual-quaternions, i.e., a \textit{pure rotation} and \textit{pure translation}:
\begin{equation} \label{eq:pure_dual_quats}
\begin{alignedat}{3}
	\underline{\hat{q}}_{tra} &= 1       + \epsilon \frac{1}{2} \hat{q}_{pos} && \quad \text{(pure translation)}\\
	\underline{\hat{q}}_{rot} &= \hat{q}_{rot} + \epsilon \hat{0} && \quad \text{(pure rotation)}\\
\end{alignedat}
\end{equation}

\noindent where we concatenate the pure dual-quaternion transforms by multiplication to calculate the combined set of transforms, as shown in Equation (\ref{eq:dual_quat_parts}); however, be warned the order of multiplication determines if translation or rotation is performed first.  For example, we represent the FK problem as:

\begin{equation}
\begin{alignedat}{3}
	\underline{\hat{q}} &= 	\underline{\hat{q}}_0
							\underline{\hat{q}}_1
							\underline{\hat{q}}_2
							...
							\underline{\hat{q}}_{n-1}	\\
						&=  (\underline{\hat{q}}_{0}^{t}  \underline{\hat{q}}_{0}^{r})
							(\underline{\hat{q}}_{1}^{t}  \underline{\hat{q}}_{1}^{r})
							(\underline{\hat{q}}_{2}^{t}  \underline{\hat{q}}_{2}^{r})
							...
							(\underline{\hat{q}}_{n-1}^{t} \underline{\hat{q}}_{n-1}^{r})
\end{alignedat}
\end{equation}

\noindent where the superscript letter $r$ and $t$ indicates a pure rotation or pure translation dual-quaternion transform respectively (as defined in Equation (\ref{eq:pure_dual_quats}) ).


\section{Jacobian Matrix} \label{sec:jacobian_matrix}
\begin{figure}[!b]
	\begin{center}
	\includegraphics[width=0.55\columnwidth]{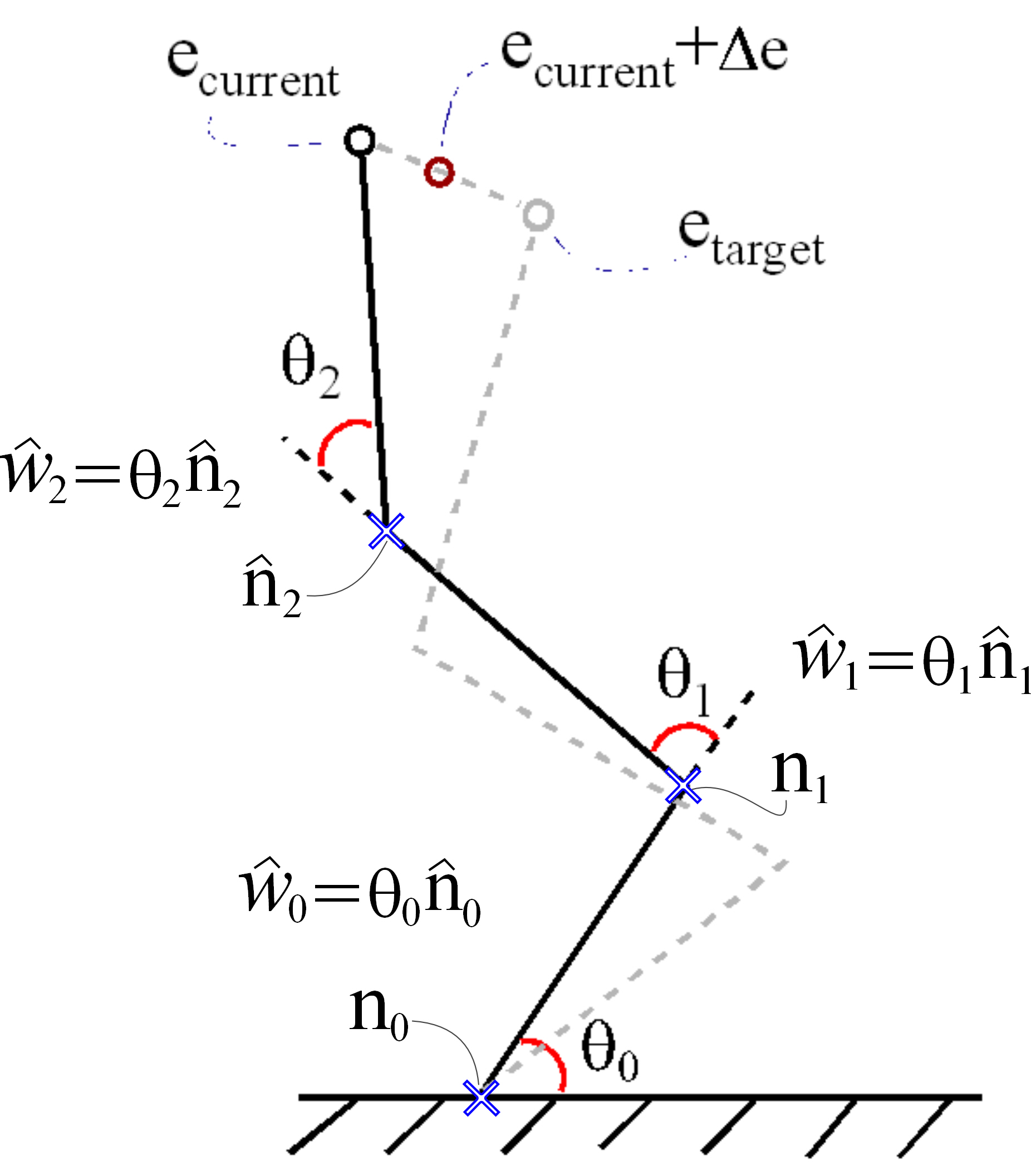}
	\end{center}
	\caption{\text{Forward \& Inverse Kinematics Example}.  Illustrating the relationship between the different parameters, e.g., end-effector error and joint orientations.  $\vec{w}$ quaternion exponential-map for each joint (i.e., axis-angle combination $\vec{w}=\hat{n}\theta$).}
	\label{fig:example_links}
\end{figure}

The Jacobian \textbf{J} is a matrix that represents the change in joint orientations to displacement of end-effectors.  Each frame we calculate the Jacobian matrix from the current angles and end-effectors.  We assume a right-handed coordinate system.  To illustrate how we calculate the Jacobian for an articulated system, we consider the simple example shown in Figure \ref{fig:example_links}.  For a more detailed description see \cite{Balestrino1984,Wolovich1984,Baillieul1985,Wampler1986,Nakamura1986,Buss2004}.  The example demonstrates how we decompose the problem and represent it as a matrix for a sole linked chain with a single three DOF end-effector.  We then extend this method to multiple linked-chains with multiple end-effectors (each with six DOF) to represent the character hierarchy.

Each joint is stored as an axis-angle component ($w=\hat{n}\theta$):
\begin{align}
\begin{split}
{\textbf{w}} = \left[ \begin{array}{l}
{\vec{w} _0}\\
{\vec{w} _1}\\
{\vec{w} _2}\\
...\\
{\vec{w} _n}
\end{array} \right]
\end{split}
\end{align}

\noindent where $\vec{w}_i$ is the rotation (i.e., axis-angle) of joint $i$ relative to joint $i-1$, and $e$ for the end-effectors global position.  From these matrices, we can determine that the end-effectors, and the joint angles are related.  This leads to the forward kinematics definition, defined as:

\begin{align}
\begin{split}
{\textbf{e}} = \left[ \begin{array}{l}
{e_x}\\
{e_y}\\
{e_z}
\end{array} \right]
\end{split}
\end{align}

The end-effectors and the joint orientation (i.e., quaternion-exponent) are related and is defined by:
\begin{equation}
\textbf{e} = f({\textbf{ w }})
\end{equation}

We can differentiate the kinematic equation for the relationship between end-effectors and joint orientation.  This relationship between a change in joint orientation and a change in end-effectors location is represented by the Jacobian matrix and is given by:
\begin{equation}
{\bf{\dot e}} = {\bf{J\dot w }}
\end{equation}

The Jacobian $\textbf{J}$ is the partial derivatives for the change in end-effectors locations by change in joint angles.

\begin{equation}
{\bf{J}} = \frac{{\partial {\bf{e}}}}{{\partial {\bf{w }}}}
\end{equation}

If we re-arrange the kinematic problem: 

\begin{equation}
{\bf{ w }} = {f^{ - 1}}({\bf{e}})
\end{equation}

We can conclude a similar relationship for the Jacobian:
\begin{equation}
{\bf{\dot w }} = {{\bf{J}}^{ - 1}}{\bf{\dot e}}
\end{equation}

For small changes, we can approximate the differentials by their equivalent deltas:
\begin{equation}
\Delta {\bf{e}} = {{\bf{e}}_{target}} - {{\bf{e}}_{current}}
\end{equation}

For these small changes, we can then use the Jacobian to represent an approximate relationship between the changes of the end-effectors with the changes of the joint angles. 

\begin{equation}
\Delta {\bf{ w }} = {{\bf{J}}^{ - 1}}\Delta {\bf{e}}
\end{equation}

We can substitute the result back in:
\begin{equation}
{{\bf{ w }}_{current}} = {{\bf{ w }}_{previous}} + \Delta {\bf{ w }}
\end{equation}

\begin{figure}
	\begin{center}
	\includegraphics[width=0.8\columnwidth]{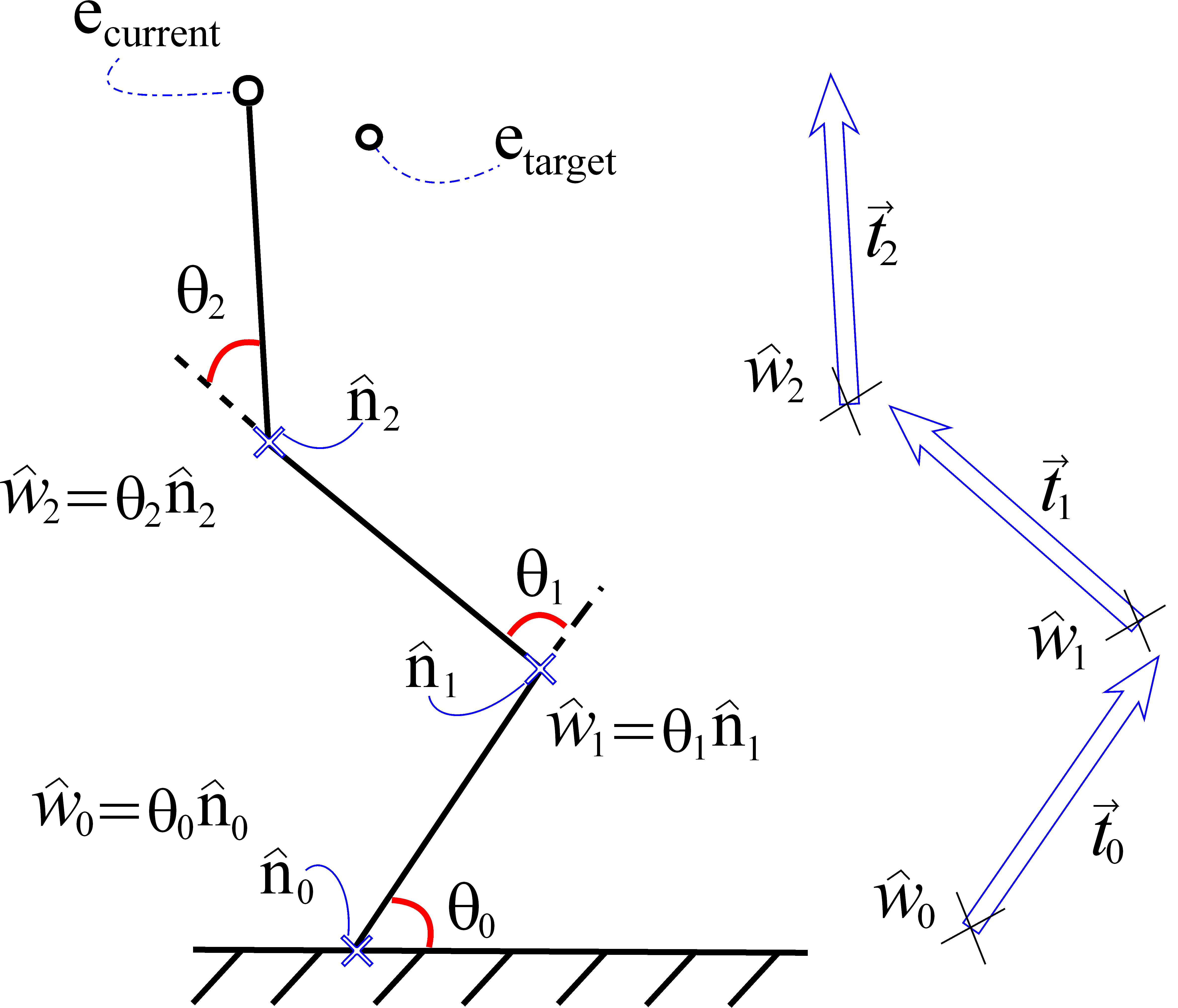}
	\end{center}
	\caption{A simple three link serial chain example, where $\vec{w}$ is the axis-angle combination (i.e., quaternion- exponential-map), $\vec{t}$ is the translation vector, $\hat{n}$ is the unit-length axis of rotation, and $\theta$ is the angle.  For a simple 2D case (i.e., only in the x-y axis) the axis of rotation is $\hat{n}=[0,0,1]$..}
	\label{fig:example}
\end{figure}

For a step-by-step explanation of the process of calculating the Jacobian for the quaternion-exponent see Appendix \ref{sec:jac_exp_how}.  For example, calculating the Jacobian for Figure \ref{fig:example} gives:
\begin{equation}
\begin{alignedat}{3}
{\bf{J}} &= \left[ {\begin{array}{*{20}{c}}
{\frac{{\partial {\bf{e}}}}{{\partial {w _0}}}}\\
{\frac{{\partial {\bf{e}}}}{{\partial {w _1}}}}\\
{\frac{{\partial {\bf{e}}}}{{\partial {w _2}}}}
\end{array}} \right]\\
\text{and}\\
{\bf{e}} &= {{\bf{e}}_{current}} - {{\bf{e}}_{target}}
\end{alignedat}
\end{equation}

The Jacobian matrix is calculated for the system so that we can calculate the inverse and hence the solution.  Alternatively, a good explanation of the Jacobian and its applications is also presented by Buss \cite{Buss2004}, who gives an introduction to IK methods using the Transpose, Pseudoinverse, and Damped Least-Square method.

\section{Gauss-Seidel Algorithm} \label{sec:gauss_seidel_algorithm}
We set up the IK problem into a particular arrangement, so that we can solve for the unknowns using the Gauss-Seidel method.  Whereby, we construct the IK formulation using the Jacobian matrix with the linear equation format of the form:

\begin{align}
\begin{split}
\text{Linear Equation:}\\
\textbf{A}\textbf{x} = \textbf{b}
\end{split}
\end{align}

The IK problem is then composed as: 

\begin{equation}
\textbf{J}^T \textbf{J} \Delta \textbf{q} = \textbf{J}^T \Delta \textbf{e}
\end{equation}

Equating equivalent variables:

\begin{align}
\begin{split}
\textbf{A} &= \textbf{J}^T \textbf{J}\\
\textbf{b} &= \textbf{J}^T \Delta \textbf{e}\\
\textbf{x} &= \Delta \textbf{q} = unknown
\end{split}
\end{align}


\subsection{Damping and Stability}
\subsubsection{Damping}
With the Gauss-Seidel iterative method, we solve for the unknown $x$ value.  To prevent singularities and make the final method more stable and robust we incorporated a damping value:

\begin{equation}
\textbf{A} = (\textbf{J}^T \textbf{J} + \delta \textbf{I})
\end{equation}

\noindent where $\delta$ is a damping constant (e.g.,  $\mathtt{\sim}$0.0001), and $\textbf{I}$ is an identity matrix.

\subsubsection{Singularities}
The exponential-map of a quaternion (i.e., the axis-angle combination) is parameterized in $\in \mathbb{R}^3$ and hence contains singularities similar to Euler angles possessing gimbals lock singularities.  However, for our IK iterative situation, we take small incremental steps (i.e., angular change is less than $\pi$) we can avoid the singularity problem, since we can shift the exponential-map singularity away from the safe working region \cite{Gras98}.  The exponenential-map has singularities at a radius of $2n\pi$ (for $n=1,2,3..$).  Hence, if the exponential-map angle $||\vec{w}||$ is close to $\pi$ we replace $\vec{w}$ by $(1-2\pi/||\vec{w}||)\vec{w}$, which is the same rotation but shifted away from the singularity problem.


\subsection{Gauss-Seidel Implementation}

The Gauss-Seidel iterative algorithm is a technique developed for solving a set of linear equations of the form $Ax=b$.  The method has gained a great deal of acclaim in the physics-based community for providing a computationally fast robust method for solving multiple constraint rigid body problems \cite{Jourdan1998,Liu2005,Catto2005}.  The iterative algorithm is based on matrix splitting \cite{Gene1996}, and its computational cost per iteration is O(n), where n is the number of constraints.  Furthermore, the number of constraints and the number of iterations is what dominates the performance of the algorithm.  Algorithm \ref{alg:gauss-seidel} is the basic Gauss-Seidel method for a generic linear system of equations of the form $Ax=b$; for the unknowns, an initial guess  is needed.  Naively this value could be zero and result in the system having a cold start.  Then the algorithm would proceed, while at each iteration, the corresponding elements from $A$, $b$ and $x$ (current) act as a feedback term to move $x$ (next) closer to the solution.

\begin{algorithm} 
\caption{Gauss-Seidel iterative algorithm to solve $Ax=b$ given $x^0$}
\label{alg:gauss-seidel}
\begin{algorithmic} [1]
    \State $x=x^0$
    \For {iter=1 to iteration limit}
    	\For {i=1 to n}
    	
    		\State $\Delta {x_i} = \frac{{\left[ {{b_i} - \sum\limits_{j = 1}^n {{A_{ij}}{x_j}} } \right]}}{{{A_{ii}}}}$
    		
    		\State ${x_i} = {x_i} + \Delta {x_i}$

    	\EndFor
	\EndFor
\end{algorithmic}
\end{algorithm}

The conditions for the Gauss-Seidel iterative Algorithm~\ref{alg:gauss-seidel} terminating are:
\begin{itemize}
	\item If a maximum number of iterations has been reached
	\item If the error $||Ax-b||$ drops below a minimum threshold
	\item If $|| \Delta x_i ||$ falls below a tolerance
	\item If $|| \Delta x_i ||$ remains the same as the previous frame (within some tolerance)
\end{itemize}

It is essential that the coefficients along the diagonal part of the matrix be dominant for the Gauss-Seidel method to converge on a solution.

\subsection{Angular Limits - Twist-and-Swing}

Any practical character-based IK system needs to have the ability to enforce angular joint limits before it can be considered a viable real-world solution.  We incorporate angular joint limits into the simple iterative algorithm by clamping the modified angle orientations at each iteration update (see Equation (\ref{eq:iteration_clamp})).  While this can be accomplished easily with Euler angles by setting a minimum and maximum angle.  For the exponential-map (i.e., the axis-angle combination) parameterization, we use the twist-and-swing decomposition~\cite{Gras98}, since it presented a fast, robust, and simple technique for robustly calculating angular errors and enforcing limits (as demonstrated and shown by Kallmann~\cite{Kall07}).

\begin{equation} \label{eq:iteration_clamp}
\begin{alignedat}{3}
\textbf{w}  = \left\{ \begin{array}{l}
lower\,\,\,\,\,\,\,\,\,\,\,\,\,\,\,\,:\,\,\,if\,(\textbf{w}  + {\textbf{J}^{ - 1}}\Delta \textbf{e})\,\, < lower\\
upper\,\,\,\,\,\,\,\,\,\,\,\,\,\,\,\,:\,\,if\,(\textbf{w}  + {\textbf{J}^{ - 1}}\Delta \textbf{e})\,\, > upper\\
\textbf{w}  + {\textbf{J}^{ - 1}}\Delta \textbf{e}\,\,\,:\,\,otherwise
\end{array} \right.
\end{alignedat}
\end{equation}

For complex joint models, such as the ball-and-socket joint, the twist-and-swing decomposition presents a practical and intuitive representation.  The twist-and-swing allows us to define and enforce joint limits intuitively.  The twist is around the `x-axis' while the swing is around the `yz-plane'.  We can decompose a quaternion orientation into its twist and swing components shown in Equation~\ref{eq:twist_and_swing}.  This is in world space but can easily be converted to local space (e.g., joint space).

\begin{align} \label{eq:twist_and_swing}
\begin{split}
{{\bf{q}}_{twist_x}} &= \left( {\frac{{{q_s}}}{{\sqrt {q_s^2 + q_x^2} }},\frac{{{q_x}}}{{\sqrt {q_s^2 + q_x^2} }},0,0} \right)\\
{{\bf{q}}_{swing_{yz}}} &= \left( {\sqrt {q_s^2 + q_x^2} ,0,\frac{{{q_s}{q_y} - {q_x}{q_z}}}{{\sqrt {q_s^2 + q_x^2} }},\frac{{{q_s}{q_z} + {q_x}{q_y}}}{{\sqrt {q_s^2 + q_x^2} }}} \right) \\
{\bf{q}} &= {\bf{q}}_{swing_{yz}}  {\bf{q}}_{twist_x}
\end{split}
\end{align}

\begin{align} \label{eq:twist_and_swing_y}
\begin{split}
{\bf{q}}_{twist_y} &= 
\left( 
\frac{q_s}{\sqrt {q_s^2 + q_y^2} },
0,
\frac{{{q_y}}}{\sqrt {q_s^2 + q_y^2} },
0
\right) \\
{\bf{q}}_{swing_{xz}} &= 
\left( 
\sqrt {q_s^2 + q_y^2},
0,
\frac{q_s q_x + q_y q_z}{\sqrt {q_s^2 + q_y^2} },
\frac{q_s q_z - q_x q_y}{\sqrt {q_s^2 + q_y^2} } 
\right) \\
{\bf{q}} &= {\bf{q}}_{swing_{xz}}  {\bf{q}}_{twist_y}
\end{split}
\end{align}

\begin{align} \label{eq:twist_and_swing_z}
\begin{split}
{\bf{q}}_{twist_z} &= 
\left( 
\frac{q_s}{\sqrt {q_s^2 + q_z^2} },
0,
0,
\frac{{{q_z}}}{\sqrt {q_s^2 + q_z^2} }
\right) \\
{\bf{q}}_{swing_{xy}} &= 
\left( 
\sqrt {q_s^2 + q_z^2},
0,
\frac{q_s q_x - q_y q_z}{\sqrt {q_s^2 + q_z^2} },
\frac{q_s q_y - q_x q_z}{\sqrt {q_s^2 + q_z^2} } 
\right) \\
{\bf{q}} &= {\bf{q}}_{swing_{xz}}  {\bf{q}}_{twist_y}
\end{split}
\end{align}

\noindent where $q_x$, $q_y$, and $q_z$ are the rotations around the x-, y-, and z-axis respectively, and $q_{xy}$, $q_{xz}$, and $q_{yz}$ are the rotations a vector in the xy-, xz-, and yz-plane respectively (see Appendix~\ref{sec:twist_swing_proof} for proof).  We can validate the twist-and-swing decomposition by multiplying them together and reconstructing the original quaternion.

This extension of the basic Gauss-Seidel algorithm to handle constraint limits for the unknowns is called the Projected Gauss-Seidel (PGS) algorithm.  The angular limits form bounds that are in form of upper and lower joint angles that are easily enforced through clamping.  Furthermore, the PGS algorithm has O(n) running time and convergence is guaranteed as long as the matrix is positive definite \cite{Cottle1992}.  In practice, we have found the algorithm to provide promising visual and numerical results.

\section{Spatial and Temporal Coherency}
We give the iterative solver a warm-start approximation at the start of each iteration update by taking advantage of spatial and temporal coherency of the problem.  Since the PGS solver is iterative by design and without a warm-start approximation, its convergence rate can be very slow (i.e., depending upon the eigenvalues of the matrix it is solving).  However, by caching the result between updates (i.e., use previous solution as the start for the next update), we can considerably reduce the number of iterations, especially for cases when there are only minuscule changes for the system.

\begin{figure}
	\begin{center}
	\includegraphics[width=1.0\columnwidth]{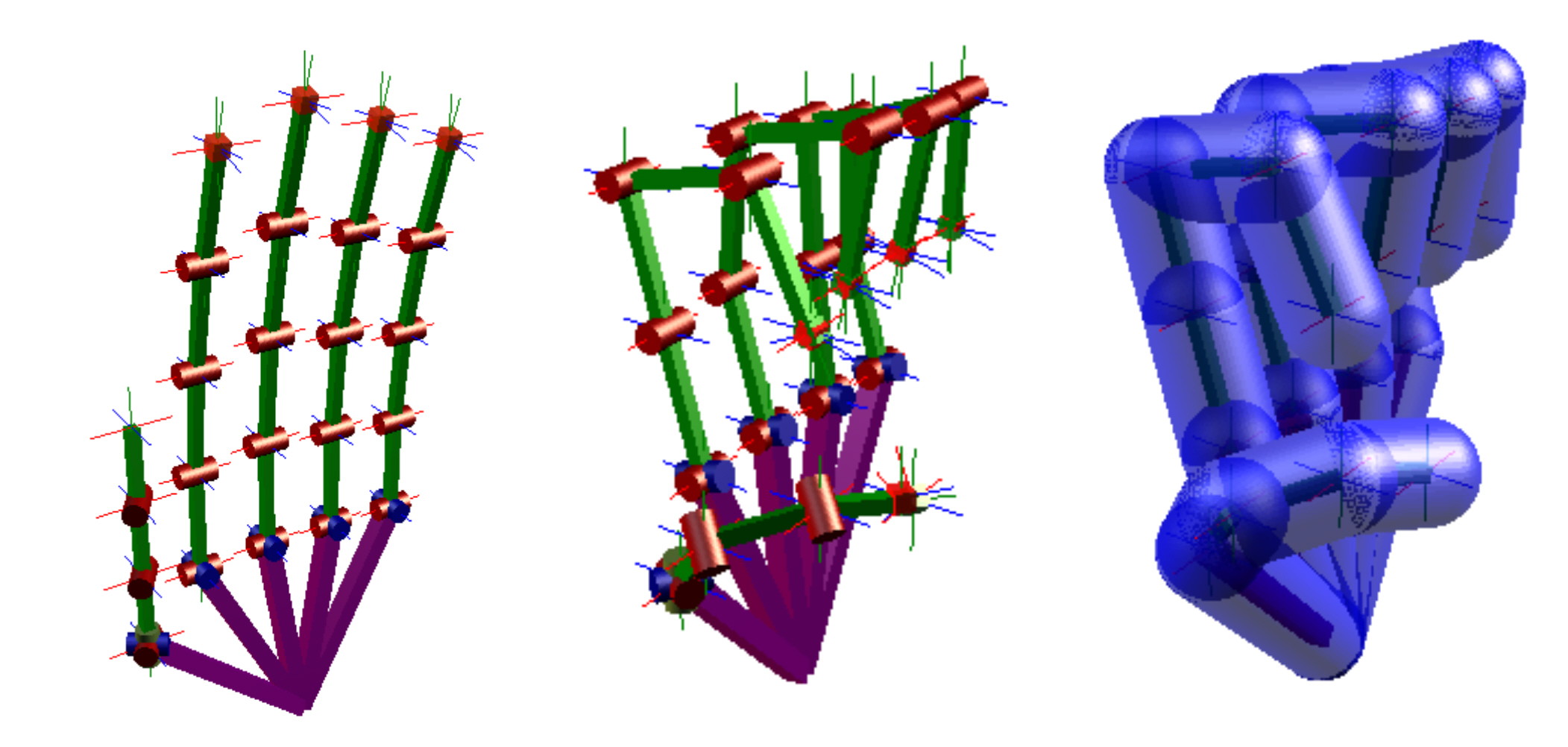}
	\end{center}
	\caption{\text{Simulation Screenshots A}.  Experimental poses for our articulated hand.}
	\label{fig:sim_renders}
\end{figure}

\section{Experimental Results} \label{sec:experimental_results}

On average, the small spatial coherent transitions between frame updates resulted in the Gauss-Seidel method requiring only two or three iterations for the end-effectors to reach acceptable answers.  This resulted in the IK solver being able to easily maintain a low-computational overhead and run at real-time frame-rates.  While our Gauss-Seidel implementation was straightforward it was implemented in a single threaded program and did not exploit any parallel architecture speed ups (e.g., using a multi-core CPU or GPU); however, numerous methods were demonstrated by Courtecuisse et al. \cite{Courtecuisse2009} for exploiting multi-core architectures to achieve much improved performance using the Gauss-Seidel algorithm.
 
The performance of our iterative Gauss-Seidel IK implementation was computationally fast and ran at real-time frame-rates enabling the IK problem to be modified on the fly.  For cases when little or no movement occurred the solver would perform 1 to 2 iterations at most, while for sporadic changes in the articulated posture resulted in approx. 10 or more iterations.  Furthermore, our Gauss-Seidel method would only require a few milliseconds to compute the solution.  However, the cost of calculating the IK solution can vary greatly depending upon the starting assumption.  Our implementation performed at real-time rates and maintained a consistent frame-rate well above a 100Hz.  Simulations were performed on a machine with the following specifications: Windows7 64-bit, 16Gb Memory, Intel i7-2600 3.4Ghz CPU.  Compiled and tested with Visual Studio 2012.

One important criteria was that the IK solver remained stable, e.g., when the end-effectors are placed out of reach, so that no solution exists.  In practice, when no result was obtainable, a best reach condition was always presented, stretching to obtain the end-effectors but remaining stable (i.e., not oscillating or jittering).  Furthermore, when end-effectors were started at radically different locations, the resulting solution would haphazardly jerk; however, the result always converged on acceptable poses.

We experimented with a diverse range of poses of generally unpredictable and chaotic finger movement to explore the stability and flexibility of our approach.  For example, we did random on the spot compositions of the hand opening, closing, stretching, and so on, and always converged on a solution.  It should be pointed out that the hand has less problems with angular limits and singularity conditions compared to a full body articulated structure.  However, the hand has enough degrees of freedom and flexibility to perform a suitably detailed set of tests.

\begin{figure}
	\begin{center}
	\includegraphics[width=1.0\columnwidth]{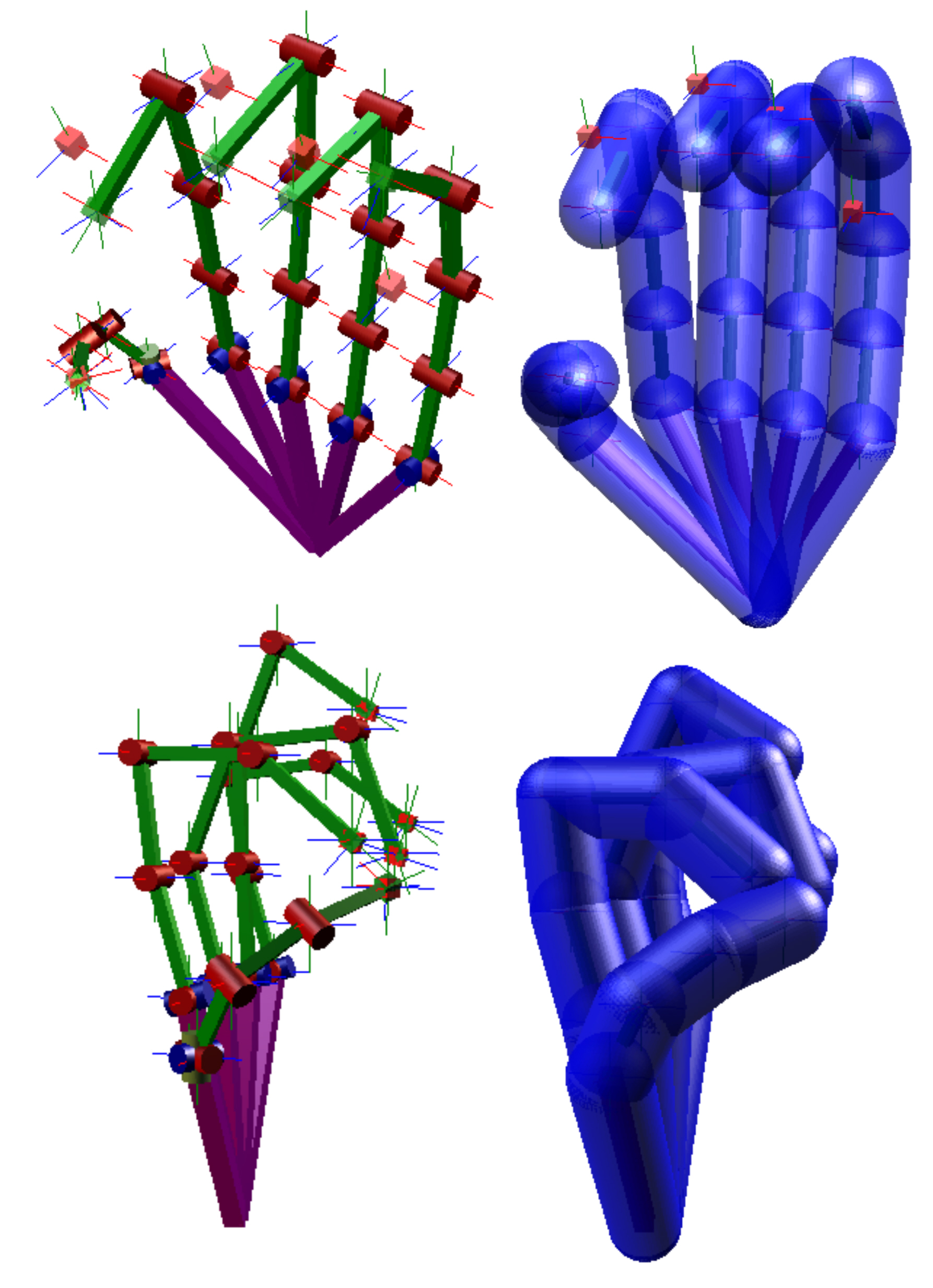}
	\end{center}
	\caption{\text{Simulation Screenshots B}.  Experimental poses for our articulated hand.}
	\label{fig:sim_renders_B}
\end{figure}

\section{Limitations}  \label{sec:limitations}
We did not include any inter-link collision detection so it was possible for fingers to pass through one another.  During situations when joints were against their angular limits, it would take longer for the iterative IK solver to converge on a solution, since joints that could not move further would be constantly pushed back.  Finally, we did not weight or couple any of the angular joints; hence, the final pose could look uncomfortable and unnatural while still being physically-plausible.  For example, in a real-world human hand, if the index finger is pulled downwards towards the wrist, it should affect its neighboring fingers.

\section{Conclusion and Discussion} \label{sec:conclusion_and_discussion}
We presented the Gauss-Seidel technique as a method for solving real-time articulated IK problems with quaternion-exponential maps and dual-quaternions.  We used temporal caching to reduce the computational cost and gain real-time performance speeds.  The results of the IK system performed well enough to be used in time critical systems (such as games).  With the angular limits, the method can suffer from singularity problems if the end-effectors jump; however, due to the end-effectors following small spatial transitions singularities are mostly avoided.  All in all, the algorithm is simple to implement, computationally fast, little memory overhead, and is fairly robust.  The IK solution can work with multiple end-effectors to produce poses with smooth movement with and without constraints.

While we demonstrated the practical aspect of using the Gauss-Seidel method as a valid real-time method for a articulated IK system, further work still needs to be done for a more detailed statistical comparison between the aforementioned IK solutions; comparing memory, complexity and computational costs.

A further area of study would be combining the IK solver with a physics-based system (i.e., rigid body constraint solver) and explore object interaction (e.g., picking up a ball or a pencil).  Furthermore, to enable greater simulation speeds, the possible investigation and exploration of making the solver more parallel, for example, Poulsen~\cite{Poul10} demonstrated a Parallel Projected Gauss-Seidel Method.

This paper exploited the Gauss-Seidel iterative method in conjunction with a set of highly non-linear equations to solve an inverse kinematic problem for an articulated structure.  While we demonstrated the practical viability of the Gauss-Seidel method with exponential-maps, we did not implement and compare our approach with the many different numerical techniques(e.g., Newton or Broyden approach) and is an area of further investigation.


\begin{figure}
	\begin{center}
	\includegraphics[width=1.0\columnwidth]{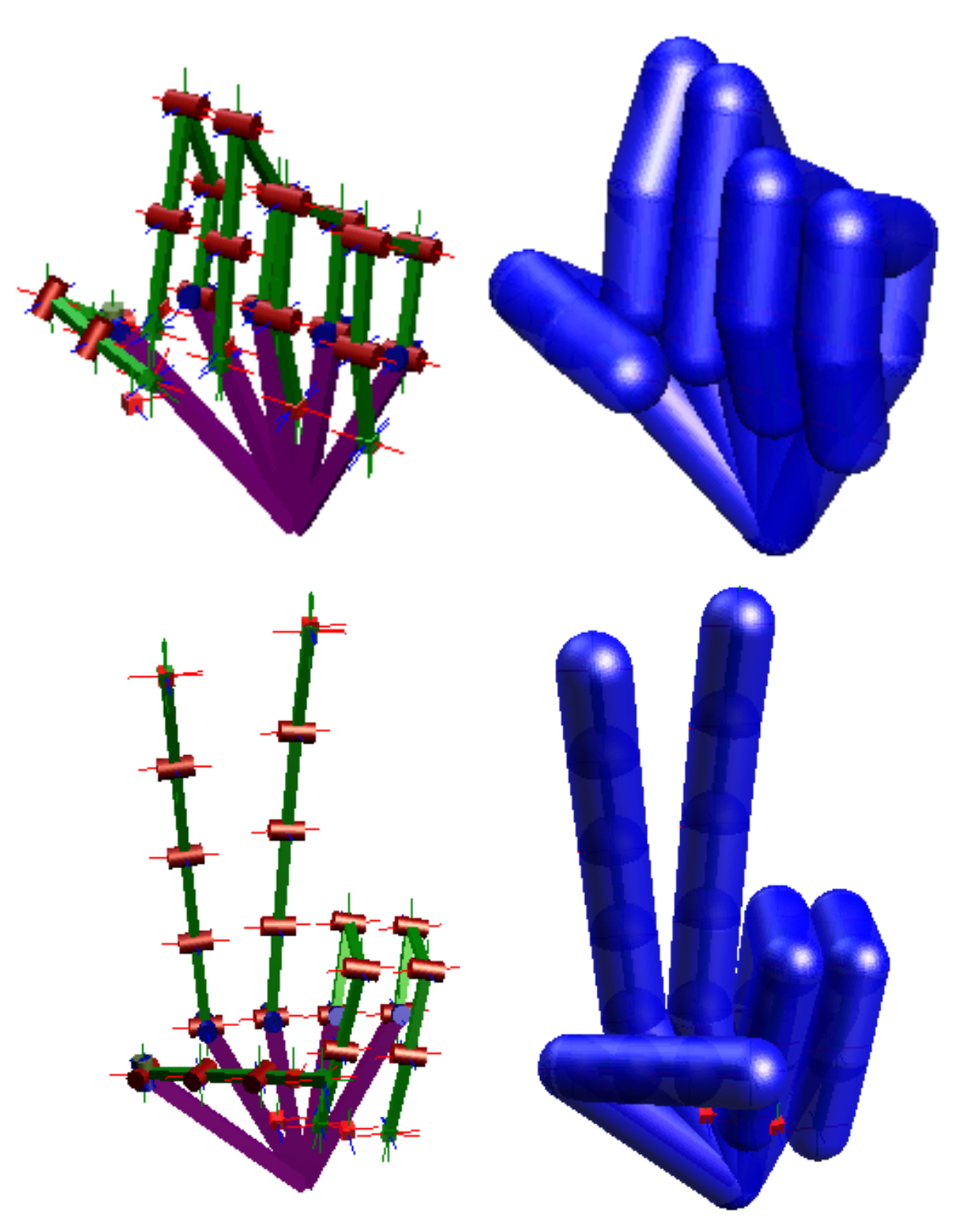}
	\end{center}
	\caption{\text{Simulation Screenshots C}.  Experimental poses for our articulated hand.}
	\label{fig:sim_renders_C}
\end{figure}

\section*{Acknowledgment}
The authors would like to thank the reviewers for taking time out of their busy schedules to provide valuable, helpful, and insightful feedback that has contributed towards making this better paper more clear, concise, and correct.



%



\bibliographystyle{ieeetr}

\bibliography{paper} 




\appendices

\section{Proof of Twist-and-Swing Decomposition Equation} \label{sec:twist_swing_proof}

We show through quaternion algebra the mathematical proof for Equation~\ref{eq:twist_and_swing}, which can similarly be applied to Equation (\ref{eq:twist_and_swing_y}) and Equation (\ref{eq:twist_and_swing_z}), and how a 3D unit-quaternion can be decomposed into two parts: the \textit{twist} and \textit{swing} components.
We start with a unit-quaternion rotation shown in Equation~\ref{eq:unit_quaternion}.
	  
\begin{equation} \label{eq:unit_quaternion}
{\bf{q}} = ({q_s},{q_x},{q_y},{q_z})
\end{equation}

\noindent where  is the vector component, and  is the scalar component.
We can calculate a quaternion from an axis-angle using Equation~\ref{eq:axis_angle}.
	  
\begin{align} \label{eq:axis_angle}
\begin{split}
{q_s} = {c_{xyz}} = \cos \left( {\frac{\theta }{2}} \right)\\
{q_x} = {s_x} = {v_x}\sin \left( {\frac{\theta }{2}} \right)\\
{q_y} = {s_y} = {v_y}\sin \left( {\frac{\theta }{2}} \right)\\
{q_z} = {s_z} = {v_z}\sin \left( {\frac{\theta }{2}} \right)
\end{split}
\end{align}

\noindent where v is a unit-vector representing the axis of rotation, and  is the angle of rotation.
Hence, we can say since the twist is only around the x-axis we can deduce that the yz-axis components will be zero and give us Equation~\ref{eq:twist_x}.
	  
\begin{equation} \label{eq:twist_x} 
{{\bf{q}}_{twist}} = {{\bf{q}}_x} = ({c_x},{s_x},0,0)
\end{equation}

Furthermore, we can also deduce that the swing x-axis component will be zero in the resulting quaternion as shown in Equation~\ref{eq:swing_yz}.
	  
\begin{equation} \label{eq:swing_yz} 
{{\bf{q}}_{swing}} = {{\bf{q}}_{yz}} = ({c_{yz}},0,{s_y},{s_z})
\end{equation}

\noindent where c and s represent the scalar cos and sin component of the half angles (i.e., see Equation~\ref{eq:axis_angle}).
A unit-quaternion must obey Equation~\ref{eq:unit_equation_law}.
	  
\begin{equation} \label{eq:unit_equation_law} 
q_s^2 + q_x^2 + q_y^2 + q_z^2 = 1
\end{equation}

Hence, from Equation~\ref{eq:twist_x} and Equation~\ref{eq:swing_yz} we can derive Equation~\ref{eq:7}.
	  
\begin{align} \label{eq:7} 
\begin{split}
c_x^2 + s_x^2 = 1\\
c_{yz}^2 + s_y^2 + s_z^2 = 1
\end{split}
\end{align}

Subsequently, if we multiply the individual twist and swing quaternions together we can reconstruct the original quaternion as shown in Equation~\ref{eq:reconstruct}.
	  
\begin{align} \label{eq:reconstruct} 
\begin{split}
{{\bf{q}}_{xyz}} & = {{\bf{q}}_{yz}}{{\bf{q}}_x}\\
 & = ({c_{yz}},0,{s_y},{s_z})({c_x},{s_x},0,0)\\
 & = (({c_x}{c_{yz}}),({s_x}{c_{yz}}),({c_x}{s_y} + {s_x}{s_z}),({c_x}{s_z} - {s_x}{s_y}))
\end{split}
\end{align}

Hence, from Equation~\ref{eq:twist_x} and knowing the vector sum of the two non-zero components from Equation~\ref{eq:reconstruct} sums up to one, we can derive $q_{twist}$, as shown in Equation~\ref{eq:sum}.
	  
\begin{align} \label{eq:sum} 
\begin{split}
{{\bf{q}}_{twist}} & = q_s^2 + q_x^2\\
 & = {({c_x}{c_{yz}})^2} + {({s_x}{c_{yz}})^2}\\
 & = c_{yz}^2(c_x^2 + s_x^2)\,\,\,\,\,\,\,\,\,\,\,\,\,\,\,(knowing,\,\,\,\,{\cos ^2} + {\sin ^2} = 1)\\
 & = c_{yz}^2
\end{split}
\end{align}

Therefore, we have Equation~\ref{eq:inv}. 
	  
\begin{equation} \label{eq:inv} 
{c_{yz}} = \sqrt {q_s^2 + q_x^2}
\end{equation}

We can multiply Equation~\ref{eq:reconstruct} by the inverse of Equation~\ref{eq:inv} to remove the quaternion swing component and leave the quaternion twist part (shown in Equation~\ref{eq:remove}).
	  
\begin{align} \label{eq:remove} 
\begin{split}
{{\bf{q}}_{twist}} & = {{\bf{q}}_x}\\
 & = ({c_x},{s_x},0,0)\\
 & = (({c_x}{c_{yz}}),({s_x}{c_{yz}}),0,0)\frac{1}{{{c_{yz}}}}\\
 & = ({q_s},{q_x},0,0)\frac{1}{{\sqrt {q_s^2 + q_x^2} }}
\end{split}
\end{align}

We extract the swing component by multiply the quaternion by the inverted (conjugated) twist quaternion (shown in Equation~\ref{eq:extract2}).
	  
\begin{align} \label{eq:extract2} 
\begin{split}
{{\bf{q}}_{swing}} & = {{\bf{q}}_{xyz}}{\bf{q}}_{twist}^*\\
 & = ({q_s},{q_x},{q_y},{q_z})({q_s}, - {q_x},0,0)\frac{1}{{\sqrt {q_s^2 + q_x^2} }}\\
 & = ({c_{yz}},0,{s_y},{s_z})\\
 & = ((q_s^2 + q_x^2),0,({q_s}{q_y} - {q_x}{q_z}),({q_s}{q_z} + {q_x}{q_y}))\frac{1}{{\sqrt {q_s^2 + q_x^2} }}
\end{split}
\end{align}

Whereby, Equation~\ref{eq:remove} and Equation~\ref{eq:extract2} sums-up our algebraic proof.  Similarly the twist-and-swing can be proved for the y-, and z-axis (as shown in Equation (\ref{eq:twist_and_swing_y}) and Equation (\ref{eq:twist_and_swing_z})).



\section{Calculating the Jacobian}

The joint-space and Cartesian space are the two state space variables for the kinematic system (shown in Figure \ref{fig:understanding_ik}).  The Jacobian matrix relates the changes in joint angles $\bf \theta$ with change in position or orientation $\bf X$ of some point on the connected hierarchy of links (i.e., Cartesian space) and is defined as:
\begin{equation}
	J = \frac{\partial {\bf X} }{\partial {\bf \theta } }
\end{equation}

\subsection{Finite-Difference Method}
The finite-difference method is an approximation technique given by Equation (\ref{eq:finite_diff}).  The error is proportional to the $\delta$ step and is limited by numerical accuracy (e.g., a 32-bit floating point) and the computational speed of the system.  The method is idea for situations where it is difficult or impossible to find an analytical solution for the kinematic system due to its complexity and size.  However, the method is a good solution for estimating an approximate Jacobian solution.

\begin{equation} \label{eq:finite_diff}
	\frac{dy}{dx} \approx \frac{y(x+\delta)-y(x)}{\delta}
\end{equation}

The finite-difference implementation for calculating an approximate Jacobian is given in Algorithm \ref{alg:finite_diff}. 

\begin{algorithm} 
  \caption{Finite-Different Method for Approximating the Jacobian}
  \label{alg:finite_diff}
\begin{algorithmic}
	\State	p = f(q)
	\For {$n=0$ to $n-1$}
		\State	$p_{\delta} = f(q+e_{n}\delta)$
		\State	${\bf J}_{n} = \frac{p_{\delta}-p}{\delta}$
	\EndFor 
	\State	return \textbf{J}
\end{algorithmic}
\end{algorithm}

\subsection{Analytical Formulation of the Jacobian using Dual-Quaternions and Quaternion Exponentials} \label{sec:jac_exp_how}

As step-by-step derivation of the Jacobian Dual-quaternion to Quaternion Exponential Mapping from the work by Bartelink \cite{Bart12} is presented below in Figure \ref{fig:dual_quaternion_jacobian}.

\begin{figure*}
	\begin{center}
	\includegraphics[width=1.0\textwidth]{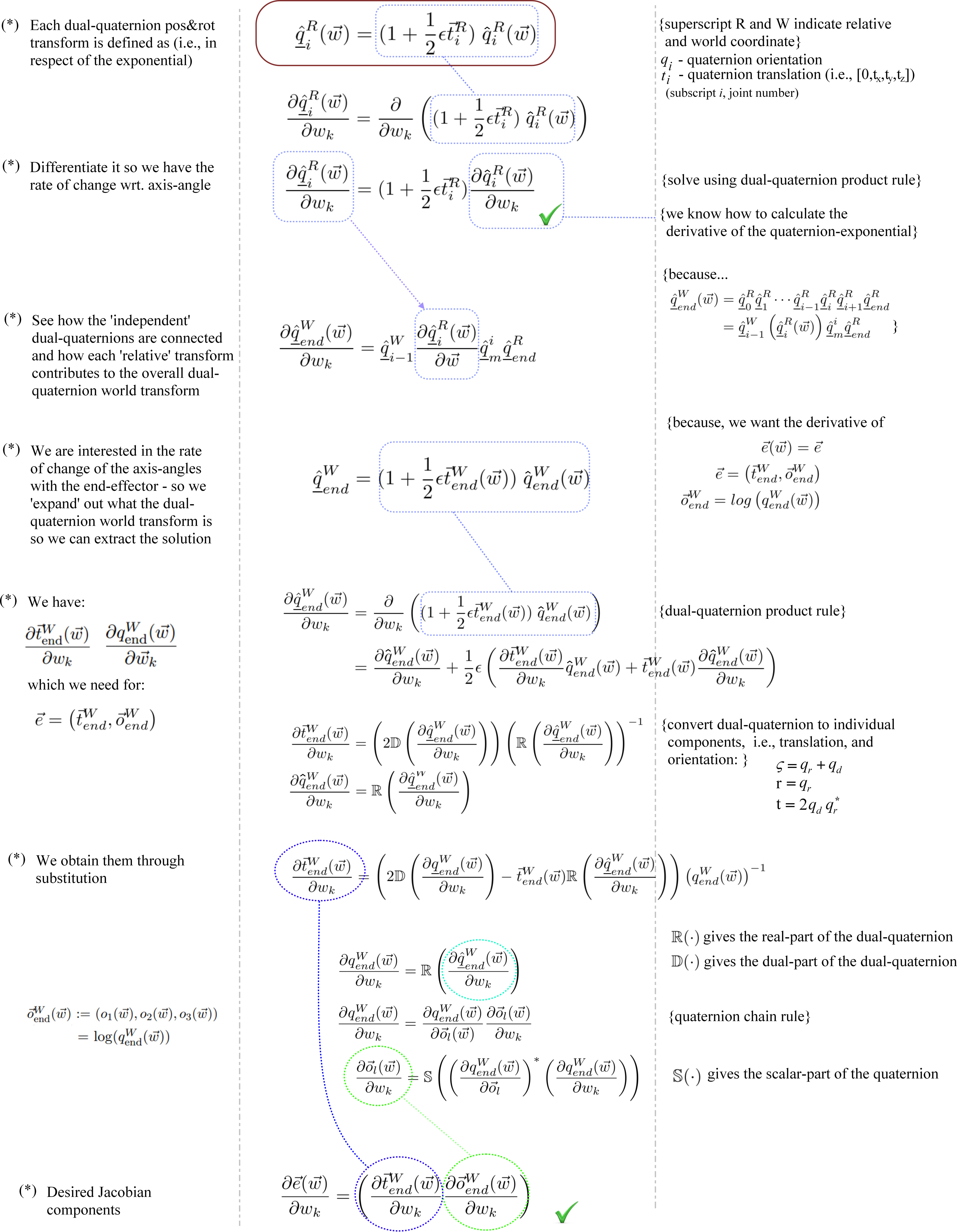}
	\end{center}
	\caption{Step-by-Step Formulation of Dual-Quaternion and Quaternion Exponential Jacobian.}
	\label{fig:dual_quaternion_jacobian}
\end{figure*}


\end{document}